\documentclass[letterpaper]{article}
\usepackage[preprint]{aaai2027}
\usepackage[hyphens]{url}
\usepackage{graphicx}
\usepackage{natbib}
\usepackage{caption}
\usepackage[utf8]{inputenc}
\usepackage{booktabs}
\usepackage{array}
\usepackage{amsmath}
\usepackage{amssymb}
\usepackage{nicefrac}
\usepackage{siunitx}
\DeclareSIUnit{\angstrom}{\text{Å}}
\title{\Huge ED-CSP: Crystal Structure Prediction \\ from Electron Diffraction}
\author{
Germain Poloudenny\textsuperscript{\rm 1,\rm 2},
Arnaud Demorti\`ere\textsuperscript{\rm 2,\rm 3,\rm 4},
Ya\"el Fr\'egier\textsuperscript{\rm 1}
}
\affiliations{
\textsuperscript{\rm 1}Laboratoire de Math\'ematiques de Lens (LML), UR 2462, Universit\'e d'Artois, Lens, France\\
\textsuperscript{\rm 2}Laboratoire de R\'eactivit\'e et de Chimie des Solides (LRCS), CNRS UMR 7314, UPJV, Amiens, France\\
\textsuperscript{\rm 3}R\'eseau sur le Stockage Electrochimique de l'Energie (RS2E), CNRS FR 3459, Amiens, France\\
\textsuperscript{\rm 4}ALISTORE-European Research Institute, CNRS FR 3104, Amiens, France
}

\begin{document}
\maketitle

\begin{abstract}
Recovering a periodic 3D crystal structure from sparse, unindexed detector-plane observations is a challenging generative inverse problem.
Prior electron diffraction (ED) learning methods largely predict crystallographic labels, reconstruct structures from indexed reflections, or retrieve structures from finite libraries.
In this paper, we consider the task of crystal structure prediction from known composition and atom count, together with multiple detector-plane ED spot sets, and introduce ED-CSP, a machine learning model that combines a relational set encoder, permutation-invariant multi-view aggregation, and a periodic flow generator to generate the lattice and fractional atomic coordinates.
To train ED-CSP, we construct Electron Diffraction Crystal Structures (ED-CS), a \num{4.85}-million-structure resource with simulated multi-view ED, deduplicated across seven materials repositories and filtered to exclude CHILI-100K matches under a fixed structural matcher.
On 2{,}075 held-out CHILI-100K materials, CHILI-only ED-CSP achieves a structural match rate (MR) of \SI[round-precision=1]{57.49}{\percent} at five candidates per query (MR@5), versus \SI[round-precision=1]{52.92}{\percent} for PXRDGen, a state-of-the-art crystal structure prediction (CSP) model conditioned on powder X-ray diffraction (PXRD); both use the same periodic-generator architecture but modality-specific encoders and inputs.
To demonstrate that increasing the dataset size improves model performance, we warm-start the full model from a separate one-million-structure precursor and show that this raises MR@5 to \SI[round-precision=1]{66.27}{\percent}.
On 1{,}024 queries whose reduced formulas are absent from the train/validation retrieval library, this registry-1M-initialized model retains \SI[round-precision=1]{53.52}{\percent} MR@5, demonstrating recovery where exact-formula lookup has no candidate.
For the same model, replacing target ED observations with those from a non-isomorphic same-formula donor on 67 queries lowers mean MR@5 by \num[round-precision=1]{22.09} percentage points across five generation seeds, providing evidence of query-specific diffraction use.
ED-CSP and ED-CS provide a controlled benchmark for generative inference from sparse simulated ED and future experimental-transfer studies.
\end{abstract}

\section{Introduction}

\begin{figure}[t]
\centering
\includegraphics[width=\columnwidth]{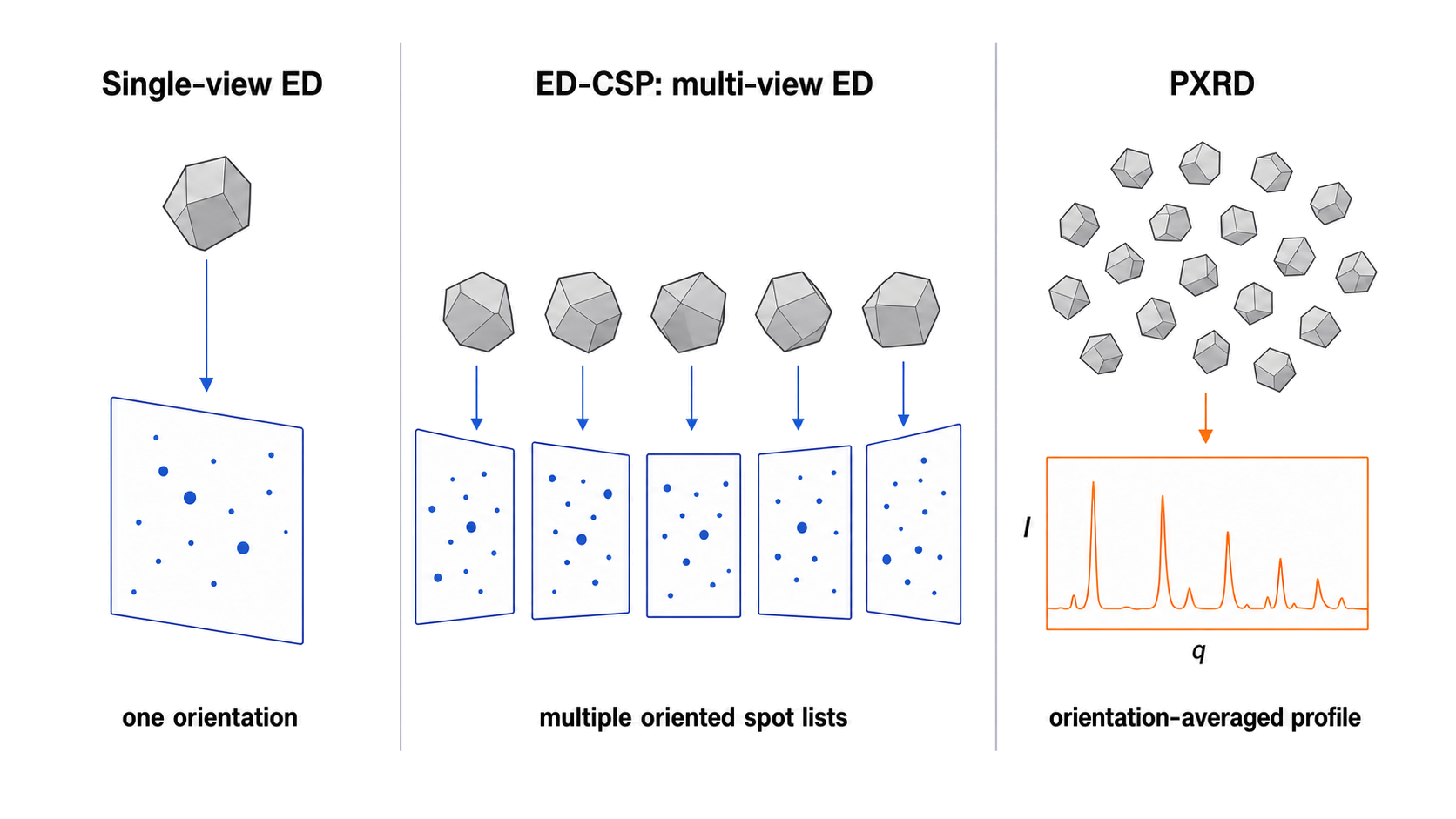}
\caption{\textbf{Diffraction observation regimes.}
ED-CSP conditions on discrete sparse ED views from multiple orientations, whereas PXRD aggregates randomly oriented crystallites into a one-dimensional radial profile.}
\label{fig:diffraction-regimes}
\end{figure}

Many crystalline materials cannot be grown as crystals large enough for conventional single-crystal X-ray diffraction, whereas ED can collect structural signal from individual nanocrystals~\cite{gemmi2019,unge2025standards}.
Powder X-ray diffraction (PXRD) remains broadly accessible for bulk powders composed of randomly oriented crystallites, making the two modalities complementary rather than interchangeable~\cite{xrdiff2026,gemmi2019}.
Unlike the orientation-averaged PXRD profile, each ED view samples an oriented region of reciprocal space and preserves detector-plane relationships among scattering vectors.
Both 3D ED and scanning diffraction experiments produce orientation-dependent reciprocal-space observations, although their acquisition geometries differ from the discrete simulated views studied here~\cite{gemmi2019,savitzky2021py4dstem}.
This orientation dependence motivates a multi-view inverse problem in which geometric information is available while simulation-to-experiment transfer and incomplete orientation coverage remain explicit limitations.
Figure~\ref{fig:diffraction-regimes} contrasts these observation regimes.

Diffraction-conditioned CSP is now established for PXRD through contrastive pretraining, diffusion and flow models, and autoregressive generation~\cite{xtalnet2024,guo2025pxrdnet,li2025pxrdgen,decifer2025,xrdiff2026}.
In ED, however, machine learning has mainly predicted crystallographic labels or retrieved structures from finite databases~\cite{gleason2024rfed,nathani2026peaggmoe,peng2026saedretrieval}.
These tasks show that sparse ED patterns contain learnable structural signal, but they stop short of predicting lattice and fractional atomic coordinates from unindexed multi-view spot lists given composition.
ED-CSP targets this gap with composition-conditioned crystal-geometry prediction.

ED-CSP encodes each view with a shared relational spot encoder, aggregates information across views, and jointly optimizes the resulting representation with a periodic flow generator.
Known composition fixes the atom types and count, while the ED branch receives detector-plane spot coordinates and intensities.

On CHILI-100K, we use a common held-out split to evaluate signal use, registry-scale transfer, finite-library coverage, same-dataset PXRD generation, and indexed-reflection reconstruction~\cite{friisjensen2024chili}.
ED input interventions and a converged composition-only reference support signal use beyond composition; registry scaling improves held-out recovery, and library-coverage stratification separates generation from finite-database lookup.
The study uses simulated detector-plane ED with known composition.

\paragraph{Contributions.}
This paper makes the following focused contributions:
\begin{itemize}
\item A formulation of composition-conditioned CSP from unindexed sparse multi-view ED, instantiated by ED-CSP.
\item Controlled ED-input interventions and a converged composition-only reference testing use of the ED conditioning signal.
\item ED-CS, a CHILI-disjoint corpus of 4{,}852{,}131 deduplicated structures with simulated multi-view ED, together with registry-scale transfer and comparisons across generation, retrieval, and indexed reconstruction.
\end{itemize}

\section{Related Work}

\paragraph{Diffraction-conditioned generation.}
PXRD-conditioned CSP has progressed from establishing diffraction as a generative condition to testing which auxiliary information is available at inference and whether diffraction resolves structural ambiguity.
XtalNet combines contrastive PXRD--structure pretraining with equivariant generation; PXRDnet and PXRDGen condition diffusion or flow generators on formula and PXRD, with PXRDGen additionally supporting lattice inference and Rietveld refinement; deCIFer generates crystallographic information file (CIF) sequences autoregressively~\cite{xtalnet2024,li2025pxrdgen,decifer2025,guo2025pxrdnet}.
PXRDGen evaluates its contrastively pretrained XRD encoders through retrieval and then uses them, either frozen or trainable, to condition structure generation~\cite{li2025pxrdgen}.
More recent systems emphasize experimental transfer and varying chemical or crystallographic inputs: XRDSol receives stoichiometry and unit-cell parameters, RealPXRD-Solver supports lattice-conditioned and lattice-free inference after large-scale simulated pretraining, and XRDiff evaluates full and partial composition with composition-grouped polymorph splits~\cite{yu2026xrdsol,realpxrd2026,xrdiff2026}.
Outside diffraction-conditioned CSP, Atomistic Language Models couple a language backbone to an atomistic diffusion decoder and report strong composition-conditioned crystal recovery, emphasizing the importance of isolating diffraction-specific gains from learned structural priors~\cite{edamadaka2026alm}.
ED-CSP addresses the complementary setting of sparse multi-view ED spot lists.

\paragraph{Sparse multi-view ED learning.}
ED representation learning provides the closest architectural precedent.
RF-ED predicts crystal systems, space groups, and lattice parameters from one or more simulated two-dimensional ED patterns, while PE-AG-GMoE processes Bragg spots as variable-size relational sets and aggregates predictions across orientations for crystallographic classification~\cite{gleason2024rfed,nathani2026peaggmoe}.
Recent multi-view selected-area electron diffraction (SAED) learning fuses two views for symmetry prediction and formula-constrained retrieval from a finite structure database~\cite{peng2026saedretrieval}.
ED-CSP changes the output from labels or database identities to lattice and fractional atomic coordinates given composition.

\paragraph{Indexed ED solution and refinement.}
Learned ED inverse methods also operate after crystallographic preprocessing.
GraPhAI phases indexed three-dimensional reflection amplitudes, while hybrid physics--ML refinement optimizes an existing structural model against integrated per-HKL intensities using differentiable dynamical simulation~\cite{melgalvis2026graphai,malik2026hybrid}.
Conventional continuous-rotation 3D ED likewise combines indexing and integration with structure solution and refinement~\cite{klar2023dynamical3ded}.
These methods complement ED-CSP but solve phasing or refinement after indexing, rather than generation from unindexed detector-plane spot lists.

\section{Method}

\subsection{Problem Setting and Inputs}

\begin{figure*}[!t]
\centering
\includegraphics[width=\textwidth]{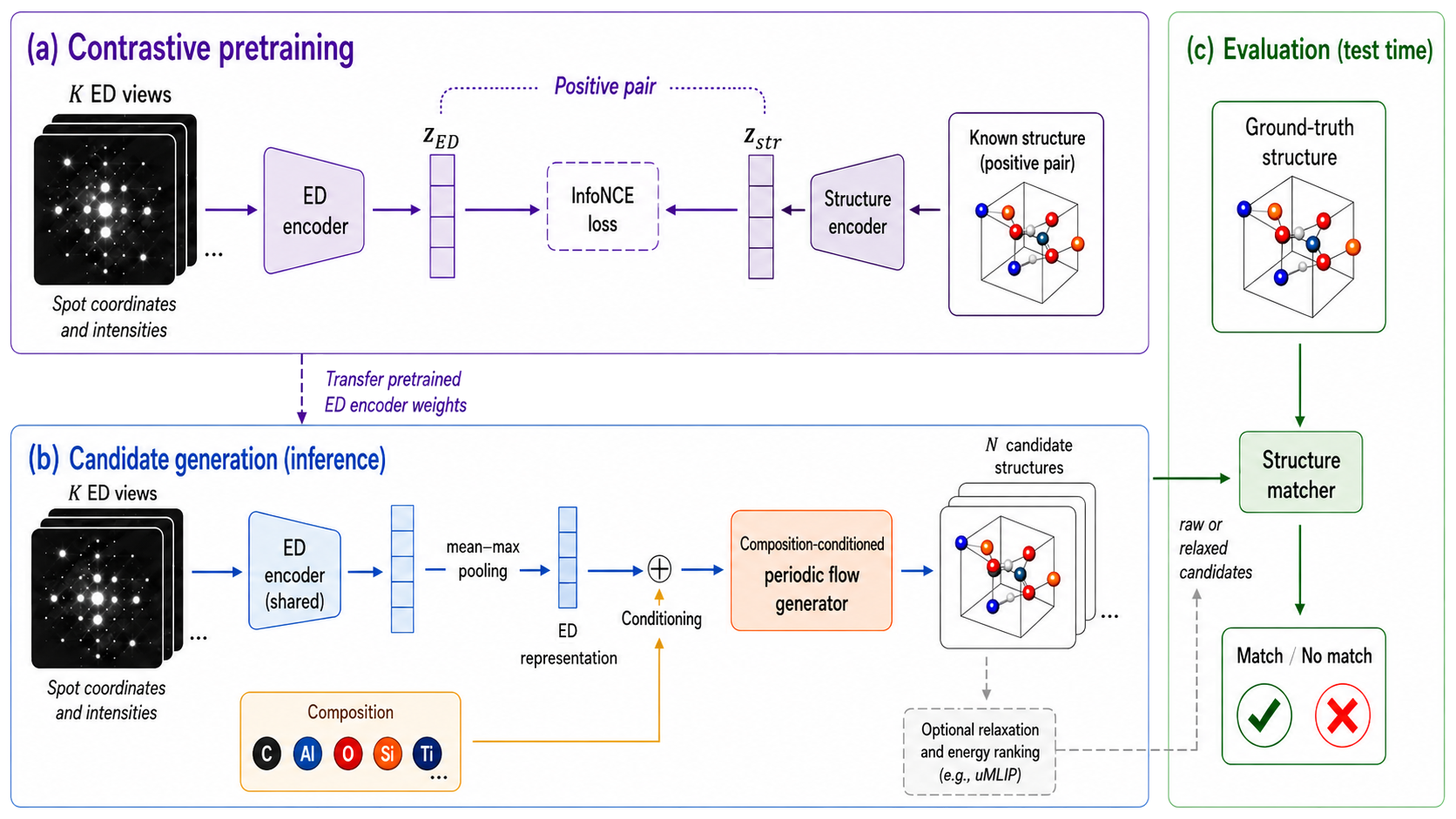}
\caption{\textbf{ED-CSP pretraining, inference, and optional post-processing.}
The optional uMLIP branch receives generated candidates but neither ED observations nor ground truth; structure matching is evaluation only.}
\label{fig:edcsp-pipeline}
\end{figure*}

Given a composition $A$ and $K$ ED views $S_{1:K}$, ED-CSP models candidate lattices $L$ and fractional coordinates $F$ through $p_\theta(L,F\mid A,S_{1:K})$.
Each view is a variable-length spot list $S_k=\{(q_x,q_y,\log(1+I))_j\}_j$, padded only at batching time.
The composition fixes the atom types and atom count, while the ED branch receives only the sampled detector-plane spot lists.
Indexed Miller labels, zone-axis vectors, and crystallographic labels are not provided to the ED branch.
Figure~\ref{fig:edcsp-pipeline} summarizes the resulting pipeline.

\subsection{Sparse Multi-View ED Encoder}

The CHILI-100K ED-CSP runs reported here use an ED encoder adapted from the EDiffCrystals PE-AG-GMoE sparse diffraction backbone~\cite{nathani2026peaggmoe}.
Each ED view is encoded by a shared PE-AG-GMoE-style graph-attention module over the raw py4DSTEM detector-plane spot list.
The resulting per-view representations are then aggregated across the sampled ED views with a mean--max pooling head and projected into the conditioning space of the generator.

\subsection{Training and Initialization}

The ED encoder is optimized jointly with the periodic generator in all reported ED-CSP runs.
For the CHILI-only checkpoint, its initial weights come from a separate ED--structure contrastive model that aligns paired multi-view observations and crystal structures in a normalized embedding space; the transferred ED encoder remains trainable and the structure encoder is discarded.
Following the PXRDGen training design, this stage serves two roles: ED-to-structure retrieval benchmarks the learned diffraction representation, and the pretrained ED weights initialize the conditioning encoder before generator training~\cite{li2025pxrdgen}.
It is particularly useful for screening encoder and view-aggregation choices: retrieval isolates the ED representation from composition conditioning and requires neither periodic-generator optimization nor iterative structure sampling.
We evaluate the first role directly; without a matched randomly initialized generator, the downstream contribution of the second is not isolated.
The registry-1M experiment instead warm-starts the complete registry-trained ED-CSP model before CHILI finetuning, so its gain measures full-model transfer rather than ED-CL alone.
This completed one-million-structure precursor predates the final ED-CS filtering and is therefore reported separately from the 4.85-million-structure corpus.

\subsection{Periodic Flow Generation}

A six-layer CSPNet-style periodic graph decoder~\cite{jiao2023diffcsp,li2025pxrdgen} conditions on flow time, the current lattice and fractional coordinates, and the aggregated ED state to predict lattice and periodic-coordinate vector fields.
Training minimizes their weighted mean-squared errors, $\mathcal{L}=\mathcal{L}_{\mathrm{lat}}+100\mathcal{L}_{\mathrm{coord}}$.
ED-CSP and PXRDGen share the same CSPFlow/CSPNet structure generator but use modality-specific encoders; their comparison therefore evaluates the complete ED- and PXRD-conditioned systems rather than isolating diffraction modality alone.

\section{Experimental Setup}

\subsection{Datasets and ED Simulation}

We use CHILI-100K, a KDD graph-ML benchmark derived from experimentally determined inorganic structures~\cite{friisjensen2024chili}.
We additionally construct Electron Diffraction Crystal Structures (ED-CS) v1, a frozen snapshot comprising 4{,}852{,}131 structures selected from AFLOW, Alexandria, the Crystallography Open Database, GNoME, Materials Project, OQMD, and JARVIS-DFT~\cite{curtarolo2012aflow,grazulis2012cod,schmidt2024alexandria,merchant2023gnome,jain2013materialsproject,saal2013oqmd,choudhary2020jarvis}.
ED-CS v1 is a curated construction snapshot, not an exhaustive mirror of any upstream repository; each source contribution is defined by its versioned eligibility, simulation, deduplication, and exclusion manifests, and later additions require a separately versioned expansion.
The v1 candidate snapshot contains entries with at most 100 sites; we merge canonical identifiers, deduplicate same-formula structures with StructureMatcher, require a certified payload with at least ten valid simulated views, and exclude every match to CHILI-100K under the same strict matcher settings.
The resulting certificate contains zero CHILI matches; source counts and the full construction record are reported in the supplementary material.
For every retained identifier, a reconstruction registry links the stored canonical structure, exact orientations, and ED arrays to record-level hashes; all 4{,}852{,}131 entries pass this technical completeness check.
The Code and Data Supplement provides the construction code, provenance and integrity schemas, and a source-stratified subset of 256 real ED-CS records; source-specific redistribution terms for the full staged corpus are detailed in the supplementary material.
Table~\ref{tab:edcs-construction} records the attrition at each certified construction stage.

\begin{table}[tb]
\centering
\small
\caption{ED-CS construction record; bold marks the final v1 snapshot.}
\label{tab:edcs-construction}
\begin{tabular}{@{}lr@{}}
\toprule
Stage & Structures \\
\midrule
Canonical entries in v1 candidate snapshot & 5{,}330{,}947 \\
After global structural deduplication & 5{,}324{,}221 \\
Entries with at least 10 certified ED views & 4{,}852{,}708 \\
\textbf{After CHILI exclusion (ED-CS)} & \textbf{4{,}852{,}131} \\
\bottomrule
\end{tabular}
\par\smallskip
ED-CS v1 is a frozen curated snapshot, not an exhaustive export of its upstream repositories. The final strict StructureMatcher certificate reports zero CHILI matches and zero matcher errors.
\end{table}

For each retained structure, we precompute dynamical ED spot patterns using the py4DSTEM simulation pipeline and parameters adopted by EDiffCrystals~\cite{savitzky2021py4dstem,gleason2024rfed,nathani2026peaggmoe}, at \SI{300}{\kilo\electronvolt} electron energy, \SI{20}{\nano\meter} thickness, and a reciprocal-space cutoff of \SI{2.0}{\per\angstrom}.
This multi-orientation protocol follows recent ML electron-diffraction benchmarks based on py4DSTEM or Bloch-wave point-list patterns~\cite{gleason2024rfed,nathani2026peaggmoe}.
Simulation starts from ten random orientations.
A throughput-oriented online policy extends a case to at most 100 views only when its ten-view pilot runtime falls below a break-even threshold estimated from recent simulation throughput and extension cost; the complete rule is given in the supplementary material.
A structure is retained only when at least ten valid views are available.
Each accepted view has at least ten spots before retaining its 16 strongest intensities.
The completed million-structure transfer curriculum uses an earlier pool drawn from Materials Project, the Crystallography Open Database, and Alexandria while retaining source provenance; it is a precursor rather than a claimed subset of the final ED-CS corpus.
The main CHILI benchmark uses 16{,}611 training structures and the same 2{,}075 ED-valid held-out queries across its reported comparisons.
ED-CSP samples ten views per structure and retains the 16 strongest spots per view.

\subsection{Training Protocol}

The CHILI-100K protocol is frozen before model evaluation.
ED-CSP is trained on the CHILI-100K train split with Adam optimization, gradient clipping, and mixed precision.
Validation match rate with one candidate selects the reported ED-CSP checkpoints, with the policy fixed before test-set evaluation.

\subsection{Baselines and Comparison Settings}

We evaluate three comparison settings with distinct inputs.
PXRDGen, XRDSol, and deCIFer provide same-dataset diffraction-conditioned generative comparisons; ED library matching measures finite-database retrieval with explicit coverage; and Superflip/EDMA measures reconstruction when simulator-indexed reflections are supplied~\cite{palatinus2007superflip,li2025pxrdgen,yu2026xrdsol,decifer2025}.
For the matched deCIFer adaptation, we train from scratch on the same 16{,}611 CHILI structures, provide only composition and a clean PXRD profile at inference, and select the checkpoint by validation loss before test evaluation.
For the XRDSol adaptation, we use the same CHILI split, its published 1{,}000-step training budget, five full 1{,}000-step diffusion samples, and its native PXRD, composition, and ground-truth unit-cell inputs.

The retrieval library is restricted to train and validation materials; test structures are excluded.
It searches all 18{,}688 ED-valid train/validation reference structures and returns five nearest neighbors.
Radial retrieval compares normalized reciprocal-radius histograms; canonical-Chamfer retrieval compares sparse two-dimensional spot sets after radial prefiltering by matching each query view to its closest library view.
Formula-aware variants restrict candidates by anonymous formula, chemical system, or exact reduced formula before ranking.
If no train/validation candidate passes a formula filter, that protocol has no valid candidate for the query.

The Palatinus-style control exports simulator-indexed reflection lists to Superflip/EDMA and uses five fixed reconstruction restarts.
It is evaluated separately from methods that receive detector-plane spot lists.

\subsection{Evaluation Protocol}

Structure generation is evaluated with the shared structural matcher implemented through pymatgen~\cite{ong2013pymatgen}, with the same matcher settings for ED-CSP, library retrieval, and all evaluations.
We report match rate (MR): the fraction of test materials for which at least one candidate structure matches the ground truth under this fixed matcher.
For CHILI-100K, ED-CSP, PXRDGen, and deCIFer are reported at both MR@1 and MR@5; XRDSol is reported at MR@5 over five independent samples; ED library matching is reported at MR@1 and MR@5 for the exact-formula control and at MR@5 for the unfiltered full-coverage control; Superflip/EDMA is reported as MR@5 over five fixed restarts.
We compute each confidence interval (CI) by non-parametric bootstrap over held-out materials and assess paired significance with a sign-flip test on the per-query ED-CSP-minus-library outcomes.

\section{Results}

Table~\ref{tab:chili-main} summarizes the CHILI-100K benchmark.
Its blocks answer different questions under a common test split and matcher; they are not an input-equivalent leaderboard.

\begin{table*}[!t]
\centering
\small
\renewcommand{\arraystretch}{1.08}
\caption{CHILI-100K benchmark on the same 2{,}075 held-out queries. Coverage is the fraction of queries for which the method has a valid input candidate set; dashes denote unmeasured metrics. Rows are grouped by their available inputs and are not a single input-equivalent leaderboard. Bold marks the strongest measured MR within the first two ED-CSP/generator blocks; control rows are not ranked.}
\label{tab:chili-main}
\begin{tabular}{@{}>{\raggedright\arraybackslash}p{3.55cm}
                >{\raggedright\arraybackslash}p{6.6cm}
                rrr
                @{}}
\toprule
Method / condition & Information available & Coverage (\%) & MR@1 $\uparrow$ (\%) & MR@5 $\uparrow$ (\%) \\
\midrule
\multicolumn{5}{@{}l}{\textit{Registry-scaled ED-CSP and signal ablations}} \\
ED-CSP + registry-1M (ours) &
py4DSTEM ED spots + composition &
100.00 &
\textbf{51.66} &
\textbf{66.27} \\
\addlinespace[1pt]
Zero ED spots &
ED spots removed (test-time intervention) &
100.00 &
17.35 &
-- \\
\midrule
\multicolumn{5}{@{}l}{\textit{CHILI-only generator comparison}} \\
ED-CSP &
py4DSTEM ED spots + composition &
100.00 &
\textbf{42.12} &
\textbf{57.49} \\
\addlinespace[1pt]
PXRDGen &
PXRD + composition &
100.00 &
34.02 &
52.92 \\
\addlinespace[1pt]
XRDSol &
PXRD + composition + ground-truth unit cell &
100.00 &
-- &
16.00 \\
\addlinespace[1pt]
deCIFer &
PXRD + composition &
100.00 &
23.28 &
34.99 \\
\midrule
\multicolumn{5}{@{}l}{\textit{Crystallographic and retrieval controls}} \\
Superflip/EDMA &
Indexed ED reflections (hidden HKL) &
99.76 &
-- &
10.51 \\
\addlinespace[1pt]
ED library, exact formula &
Train/validation ED neighbors + composition &
50.65 &
42.60 &
43.47 \\
\addlinespace[1pt]
ED library, unfiltered &
Train/validation ED neighbors &
100.00 &
-- &
18.31 \\
\bottomrule
\end{tabular}
\end{table*}

\begin{figure*}[t]
\centering
\includegraphics[width=\textwidth]{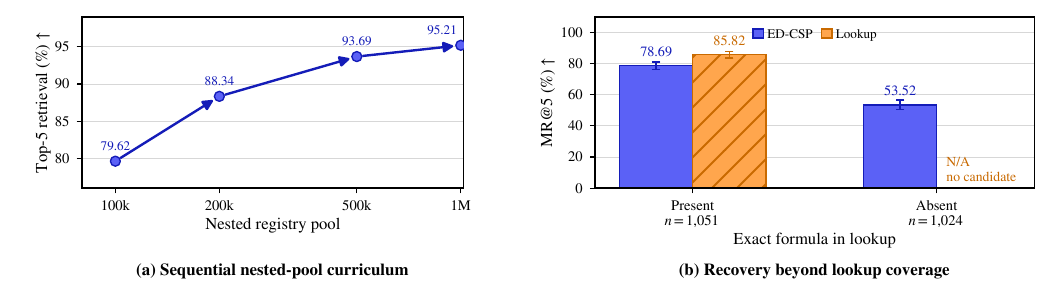}
\caption{\textbf{Effect of pretraining scale and exact-formula retrieval availability.}
(a) Sequential ED-to-structure stages: each nested-pool checkpoint resumes its converged predecessor, so the points are not independent fits.
(b) Registry-initialized ED-CSP recovery with and without an exact-formula train/validation candidate; exact-formula lookup has no candidate in the absent stratum. Error bars are query-bootstrap 95\% CIs.}
\label{fig:edcsp-results-overview}
\end{figure*}

\subsection{Registry Scaling and Signal Use}

Full-model registry-1M initialization is strongest on the aligned ED-valid split: MR@1 reaches \SI{51.66}{\percent} versus \SI{42.12}{\percent}, and MR@5 reaches \SI{66.27}{\percent} versus \SI{57.49}{\percent} for CHILI-only initialization.
The paired gains are \num{9.54} percentage points (95\% CI [\num{7.57}, \num{11.52}]) at MR@1 and \num{8.78} percentage points ([\num{7.04}, \num{10.51}]) at MR@5.
Registry-1M ED-CSP achieves \SI{66.27}{\percent} MR@5 for the designated headline seed; across three inference seeds, mean MR@5 is \SI{66.28}{\percent}, with a sample standard deviation of \num{0.12} percentage points.
We retain the reference CHILI-100K split for comparability; a post-hoc sensitivity analysis excluding 888 queries with a strict StructureMatcher near-duplicate still ranks registry-1M ED-CSP first at \SI{38.84}{\percent}/\SI{53.92}{\percent}, versus \SI{27.38}{\percent}/\SI{42.46}{\percent} for CHILI-only ED-CSP and \SI{23.67}{\percent}/\SI{38.75}{\percent} for PXRDGen.
The separate representation diagnostic in Figure~\ref{fig:edcsp-results-overview}a follows a progressive nested-pool curriculum: each scale resumes the converged checkpoint from the preceding scale, expands the training pool, and continues until validation retrieval plateaus.
Its monotonic gains show that adding registry structures improves representation retrieval along this curriculum.
Figure~\ref{fig:edcsp-results-overview}b separately evaluates full-model ED-CSP transfer.

The full-split signal ablation changes only the ED input.
Removing the ED spots reduces MR@1 from \SI{51.66}{\percent} to \SI{17.35}{\percent}, a paired drop of \num{34.31} percentage points (95\% CI [\num{32.14}, \num{36.48}]).
On the stricter 67-query subset with a non-isomorphic same-formula donor, swapping in donor ED reduces mean MR@5 by \num{22.09} percentage points across five generation seeds (95\% CI [\num{10.45}, \num{34.03}]).
A separately optimized converged composition-only reference reaches \SI{50.94}{\percent} MR@5, compared with \SI{57.49}{\percent} for CHILI-only ED-CSP on the same queries and candidate budget; because their optimization and sampling layouts differ, this \num{6.55}-point gap is descriptive rather than a paired causal estimate.
Together, the converged reference and input interventions show that ED-CSP uses ED beyond composition and learned priors on the full split, and query-specific ED on the same-formula donor subset.

\subsection{Input Sensitivity}

In a separate single-seed evaluation, we apply paired corruptions to the cached CHILI-only ED inputs while fixing the checkpoint, queries, compositions, matcher, five-candidate budget, and inference setting.
These interventions measure sensitivity to perturbed model inputs, not transfer to a new physical simulation regime.
Table~\ref{tab:chili-edcsp-ed-input-corruption-diagnostic} shows a graded response to missing spots: 10\% dropout lowers MR@5 by \num{2.07} percentage points, while 25\% lowers it by \num{7.71} percentage points.
Intensity noise at $\sigma=0.50$ produces a \num{2.02}-percentage-point decrease, whereas perturbing stored view angles by up to $5^\circ$ has no resolved effect under this cached-input protocol.

\begin{table*}[tb]
\centering
\caption{CHILI-100K sensitivity to cached ED input corruptions.}
\label{tab:chili-edcsp-ed-input-corruption-diagnostic}
\small
\begin{tabular}{lrrrr}
\toprule
ED input condition & MR@5 $\uparrow$ (\%, 95\% CI) & Matches & Queries & Delta vs. clean (95\% CI) \\
\midrule
Clean ED & 57.40 [55.28, 59.52] & 1{,}191 & 2{,}075 & 0.00 \\
10\% spot dropout & 55.33 [53.20, 57.45] & 1{,}148 & 2{,}075 & -2.07 [-3.37, -0.82] \\
25\% spot dropout & 49.69 [47.52, 51.81] & 1{,}031 & 2{,}075 & -7.71 [-9.25, -6.22] \\
Intensity noise $\sigma=0.25$ & 56.43 [54.31, 58.55] & 1{,}171 & 2{,}075 & -0.96 [-2.07, +0.14] \\
Intensity noise $\sigma=0.50$ & 55.37 [53.25, 57.49] & 1{,}149 & 2{,}075 & -2.02 [-3.28, -0.82] \\
\bottomrule
\end{tabular}
\par\smallskip
Single-seed cached-input evaluation. All deltas are paired percentage-point changes from this table's 57.40\% clean row, not from the 57.49\% primary evaluation.
\end{table*}

A separate single-seed detector-frame intervention starts from the registry-1M checkpoint and fine-tunes with random shared in-plane rotations.
It raises MR@5 under shared and independent rotations by \num{7.28} and \num{5.93} percentage points, leaving a \num{1.35}-percentage-point gap to clean inputs in both cases, at a \num{2.31}-percentage-point clean-input cost (Supplementary Material).

\subsection{Benchmark Comparisons}

The exact-formula library control reaches \SI{42.60}{\percent} MR@1 and \SI{43.47}{\percent} MR@5 at \SI{50.65}{\percent} coverage, whereas unfiltered full-coverage retrieval reaches \SI{18.31}{\percent} MR@5.
The apparent strength of formula-filtered lookup is therefore tied to analogue availability: among the 1{,}051 queries with a same-formula train/validation candidate, retrieval reaches \SI{85.82}{\percent} MR@5 versus \SI{78.69}{\percent} for registry-pretrained ED-CSP; on the remaining 1{,}024 queries it has no candidate, while ED-CSP reaches \SI{53.52}{\percent}.
This stratification separates phase lookup from out-of-library generation rather than averaging the two regimes into a misleading leaderboard.
Using exact-formula retrieval when it has coverage and ED-CSP otherwise reaches \SI{69.88}{\percent} MR@5 at the same five-candidate budget, a paired gain of \num{3.61} percentage points over ED-CSP (95\% CI [\num{2.17}, \num{5.06}]).

The indexed-reflection Superflip/EDMA control reaches \SI{10.51}{\percent} MR@5, with \SI{99.76}{\percent} valid-CIF query coverage and \SI{96.40}{\percent} valid-candidate coverage.
It consumes simulator-indexed reflections rather than detector-plane spots; its role is to measure reconstruction performance when indexed reflections are provided.

Under CHILI-only training and identical query IDs and candidate budgets, ED-CSP reaches \SI{42.12}{\percent} MR@1 and \SI{57.49}{\percent} MR@5.
PXRDGen reaches \SI{34.02}{\percent} and \SI{52.92}{\percent}, respectively.
At MR@5, ED-CSP alone solves 241 queries, PXRDGen alone solves 146, both solve 952, and both miss 736.
The matched autoregressive deCIFer adaptation reaches \SI{23.28}{\percent} MR@1 and \SI{34.99}{\percent} MR@5.
ED-CSP's paired gains over deCIFer are \num{18.84} percentage points (95\% CI [\num{16.48}, \num{21.16}]) at MR@1 and \num{22.51} percentage points ([\num{20.19}, \num{24.82}]) at MR@5; at MR@5, ED-CSP alone solves 584 queries and deCIFer alone solves 117.
XRDSol, retrained on the same CHILI split and additionally given the ground-truth unit cell, reaches \SI{16.00}{\percent} MR@5; CHILI-only ED-CSP's paired gain is \num{41.49} percentage points (95\% CI [\num{39.18}, \num{43.81}]).
These paired results compare specific systems rather than establish intrinsic ED superiority: ED-CSP receives multiple ED spot-list views through its encoder; PXRDGen uses a one-dimensional powder profile with a convolutional neural network (CNN) encoder but shares ED-CSP's CSPFlow/CSPNet generator; XRDSol additionally receives the unit cell; and deCIFer uses an autoregressive CIF-generation architecture.

\subsection{Multi-View Diagnostic}

The supplementary train-time view-count diagnostic keeps a 100-view simulation pool fixed and retrains an ED--structure retrieval encoder for each input count.
Increasing the consumed views from one to twenty improves Top-5 ED-to-structure retrieval from \SI{1.95}{\percent} to \SI{13.38}{\percent}, supporting multi-view representation learning in this encoder-level benchmark without claiming a downstream generation optimum.

\begin{figure}[ht]
\centering
\includegraphics[width=\columnwidth]{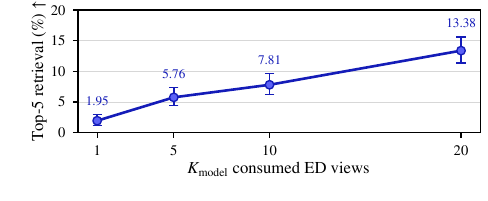}
\caption{\textbf{Train-time $K_{\mathrm{model}}$ ablation.}
Single-seed checkpoints; error bars are exact binomial 95\% CIs over 1{,}024 queries and exclude training-run variation.}
\label{fig:chili-edcsp-slice-count-ablation}
\end{figure}

\subsection{Post-Generation Relaxation}

We test whether a target-free interatomic potential can stabilize and rank a frozen five-candidate CHILI-only ED-CSP payload.
ORB-v3~\cite{rhodes2025orbv3}, MACE-MPA-0~\cite{batatia2025mace}, and eSEN-30M-OAM~\cite{fu2025esen,barrosoluque2024omat24} perform up to 100 FIRE steps, while CHGNet~\cite{deng2023chgnet} performs 30 relaxation steps; all four relax the cell and rank candidates by final energy per atom.
None of the potentials receives the ED observations or ground-truth structure.

\begin{table}[tb]
\centering
\small
\caption{Post-generation relaxation and energy ranking on CHILI-100K using a separately sampled, fixed candidate set. Deltas are paired gains over that set in percentage points; bold marks the validation-selected potential.}
\label{tab:chili-edcsp-postgen-main}
\begin{tabular}{@{}lrrrr@{}}
\toprule
Condition & Top-1 $\uparrow$ & $\Delta$ & Pool MR@5 $\uparrow$ & $\Delta$ \\
\midrule
Raw ED-CSP & 42.31 & -- & 57.06 & -- \\
\textbf{ORB-v3} & \textbf{55.47} & +13.16 & \textbf{61.69} & +4.63 \\
MACE-MPA-0 & 55.52 & +13.20 & 61.59 & +4.53 \\
eSEN-30M-OAM & 55.18 & +12.87 & 61.16 & +4.10 \\
CHGNet & 52.96 & +10.65 & 60.34 & +3.28 \\
\bottomrule
\end{tabular}
\par\smallskip
Every paired delta uses the same fixed candidate set, sampled separately from the candidates used for the 42.12\%/57.49\% primary evaluation.
\end{table}

ORB-v3 improves top-1 recovery by \num{13.16} percentage points (95\% CI [\num{11.37}, \num{14.94}]) and the relaxed candidate-pool MR@5 by \num{4.63} percentage points ([\num{3.42}, \num{5.88}]).
MACE-MPA-0 provides a near-identical independent check, with gains of \num{13.20} and \num{4.53} percentage points ([\num{11.47}, \num{14.99}] and [\num{3.33}, \num{5.73}]), respectively; ORB-v3 remains the validation-selected potential.
eSEN-30M-OAM independently gives gains of \num{12.87} and \num{4.10} percentage points ([\num{11.08}, \num{14.65}] and [\num{2.89}, \num{5.35}]), respectively, without exceeding MACE-MPA-0 or ORB-v3.
CHGNet independently gives gains of \num{10.65} and \num{3.28} percentage points ([\num{8.96}, \num{12.34}] and [\num{2.27}, \num{4.34}]), respectively.
The separately sampled frozen payload supplies its own raw baseline, which differs slightly from the designated headline evaluation.
The gains show that both candidate ordering and local geometry limit recovery, while the \SI{61.69}{\percent} relaxed-pool ceiling leaves substantial room for ED-aware refinement rather than energy-only post-processing.

\section{Discussion and Limitations}

The signal-use interventions show that the generator responds to query-specific diffraction geometry rather than treating ED as an optional auxiliary input.
Together with the coverage-stratified retrieval results, this supports generative recovery as a complement to analogue lookup and indexed-reflection workflows.

The candidate analyses identify generation quality and selection as immediate bottlenecks: relaxation and energy ranking improve top-1 recovery, yet the remaining candidate-pool ceiling indicates room for ED-aware refinement.
A natural next step is differentiable dynamical Bloch-wave refinement of generated candidates against observed ED views, jointly regularized by crystallographic or learned energy priors~\cite{malik2026hybrid}.
Beginning with indexed, orientation-aware upper bounds, this would provide an ED-specific post-generation analogue to Rietveld refinement while modeling thickness-dependent intensities.
The current ED-CSP generator consumes ten ED views, while the encoder-level diagnostic in Figure~\ref{fig:chili-edcsp-slice-count-ablation} improves Top-5 retrieval from \SI{7.81}{\percent} at $K_{\mathrm{model}}=10$ to \SI{13.38}{\percent} at $K_{\mathrm{model}}=20$, suggesting that the present input regime does not saturate multi-view representation learning.
Evaluating larger view sets in full generator training is therefore a promising direction.
ED-CS makes the scale and provenance of this direction explicit, while its simulation cost motivates adaptive allocation of orientations.
The registry results also motivate continuing to expand the unique-structure pool, which already spans millions of structures, while reducing its ED simulation cost.
Rather than assigning every structure the same simulation budget, future work should test whether K\v{r}ivovichev-style Shannon structural complexity~\cite{krivovichev2014} can help estimate the number of informative orientations required per structure.
The principal external-validity gap remains the transition from calibrated simulated point lists with known composition to experimental 3D ED, where detector calibration, uncertain spot finding, background, missing reflections, thickness variation, indexing, and expert refinement all affect the observed data.

\section{Conclusion}

ED-CSP predicts lattices and fractional atomic coordinates from known composition and sparse simulated multi-view ED.
The benchmark separates generation from finite-library retrieval and indexed-reflection preprocessing, while registry-scale full-model transfer further improves recovery.
It provides a reproducible basis for developing ED-consistent candidate ranking and refinement.

\section{Acknowledgments}
The authors gratefully acknowledge GENCI/IDRIS for providing high-performance computing resources on the Jean Zay supercomputer, which supported the computational aspects of this work. This work was also supported by MAIA (“Maitrise des Applications de l’IA”) project from alliance A2U
(Université d’Artois, UPJV et ULCO) and Hauts-de-France (HdF) region.

\bibliography{references}

\end{document}